\documentclass[letterpaper, 10 pt,journal, twoside]{ieeetran}  

\IEEEoverridecommandlockouts                              

\usepackage{graphics} 
\usepackage{epsfig} 
\usepackage{times} 
\usepackage{amsmath} 
\usepackage{amssymb}  
\usepackage{amsfonts}
\usepackage[colorlinks,linkcolor=black,anchorcolor=black,citecolor=black,urlcolor=black]{hyperref}
\usepackage[table]{xcolor}
\usepackage{booktabs}
\usepackage{multirow}
\usepackage{graphicx}
\usepackage{subfigure}
\usepackage{cite}
\usepackage{float}
\usepackage{stfloats}
\usepackage{cuted}
\usepackage{capt-of}
\usepackage{makecell}
\usepackage[ruled,vlined,linesnumbered]{algorithm2e}
\SetArgSty{textnormal} 
\makeatletter
\renewcommand{\@algocf@pre@ruled}{%
  \kern 1.5mm 
  \hrule height\algoheightrule depth0pt
  \kern\interspacetitleruled
}
\makeatother

\usepackage{etoolbox}
\makeatletter
\patchcmd{\@makecaption}
  {\scshape}
  {}
  {}
  {}
\makeatletter
\patchcmd{\@makecaption}
  {\\}
  {.\ }
  {}
  {}
\makeatother
\def\tablename{Table}

\graphicspath{{img/}}

\makeatletter
\let\@oldmaketitle\@maketitle
\renewcommand{\@maketitle}{\@oldmaketitle
    \centering  
    \setcounter{figure}{0}
    \centering
    \includegraphics[width=2.0\columnwidth]{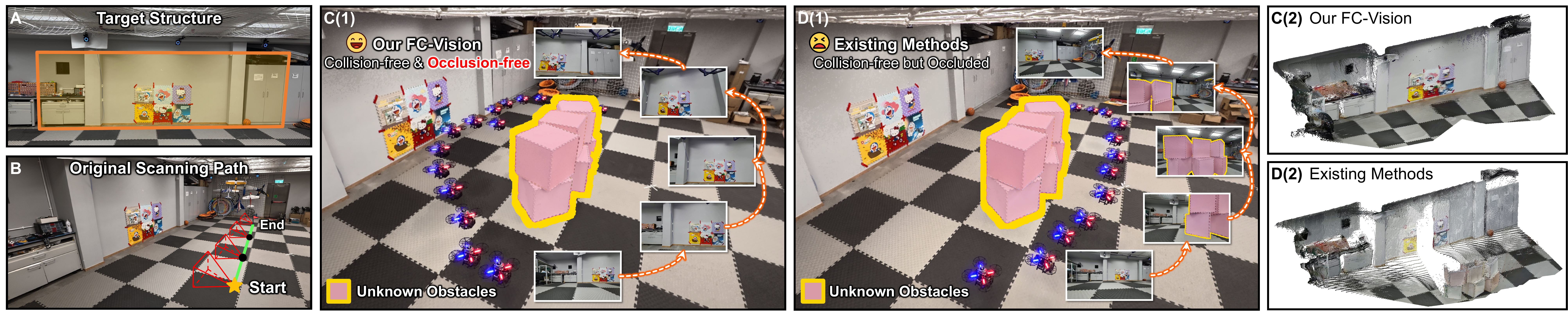}
    \vspace{-0.3cm}
    \captionof{figure}{Teaser. Given a target structure (A) and its nominal scanning path (B), \textbf{FC-Vision} enables real-time replanning to proactively avoid newly emerging unknown obstacles while enforcing collision-free and occlusion-free sensing, thereby preserving complete target coverage and efficiency of the original path (C). In contrast, existing collision-only replanning fails to prevent FoV blockage, resulting in occluded observations and degraded structural completeness (D).}
    \label{fig:Teaser}
    \vspace{-1.0cm}
}
\makeatother

\title{FC-Vision: Real-Time Visibility-Aware Replanning for Occlusion-Free Aerial Target Structure Scanning in Unknown Environments}

\author{Chen Feng$^{\dag}$, Yang Xu and Shaojie Shen
\thanks{Manuscript received: February 14, 2026; Revised: May 12, 2026; Accepted: June 15, 2026.
This paper was recommended for publication by Editor Aniket Bera upon evaluation of the Associate Editor and Reviewers' comments.}
\thanks{All authors are with the Dept. of Electronic and Computer Engineering, The Hong Kong University of Science and Technology, Hong Kong, China.}
\thanks{Email: $\{$cfengag, yxuew, eeshaojie$\}$@ust.hk, \textbf{$^{\dag}$Corresponding Author}}
\thanks{Digital Object Identifier (DOI): see top of this page.}
}

\begin{document}

\bstctlcite{IEEEexample:BSTcontrol}
\maketitle
\markboth{IEEE Robotics and Automation Letters. Preprint Version. Accepted June, 2026}{Feng \MakeLowercase{\textit{et al.}}: FC-Vision: Real-Time Visibility-Aware Replanning for Occlusion-Free Aerial Scanning}

\begin{abstract}
Autonomous aerial scanning of target structures is crucial for practical applications, requiring online adaptation to unknown obstacles during flight. 
Existing methods largely emphasize collision avoidance and efficiency, but overlook occlusion-induced visibility degradation, severely compromising scanning quality. 
This study proposes \textbf{FC-Vision}, an on-the-fly visibility-aware replanning framework that proactively and safely prevents target occlusions while preserving full target coverage and efficiency of the original plan.
Our approach explicitly enforces dense surface-visibility constraints to regularize replanning behavior in real-time via an efficient two-level decomposition: occlusion-free viewpoint repair that maintains coverage with minimal deviation from the nominal scan, followed by segment-wise clean-sensing connection in 5-DoF space. 
A plug-in integration strategy is also presented to seamlessly interface \textbf{FC-Vision} with existing UAV scanning systems without architectural changes. 
Comprehensive simulation and real-world evaluations show that \textbf{FC-Vision} consistently improves scanning quality under unexpected occluders, delivering a maximum coverage gain of 55.32\% and a 73.17\% reduction in the occlusion ratio, while achieving real-time performance with a moderate increase in flight time. 
The code has been released at \url{https://github.com/FC-Family/FC-Vision}.
\end{abstract}

\vspace{-0.2cm}

\vspace{-0.3cm}
\section{Introduction}
\label{sec:intro}

\IEEEPARstart{U}{ncrewed} aerial vehicles (UAVs) increasingly serve to autonomously scan known target structures (\textit{e.g.}, interior walls in rooms), supporting infrastructure inspection \cite{lattanzi2017review} and spatial reconstruction \cite{feng2023predrecon}.
In practice, however, surrounding environments are rarely clean or perfectly known: previously unseen obstacles, \textit{i.e.}, non-target objects that are not part of the target structure, may appear unpredictably.

A key challenge is that reactive obstacle clearance alone is insufficient for high-quality scanning.
Even when a trajectory is dynamically feasible and safe, these newly appeared obstacles may enter the sensor frustum, occluding the target and leaving unobserved surface regions.
This causes fragmented coverage and degraded structural completeness, triggering costly re-flights and increasing operational burden.
Thus, beyond safety and efficiency, a practical aerial scanning planner must adapt to unexpected objects in real-time to ensure target visibility: observations should be both complete (high coverage) and clean (suppressing non-target occlusions/clutter in the field of view (FoV)).

Existing studies address parts of this goal but rarely satisfy these requirements simultaneously.
Most model-based aerial 3D scanning planners \cite{bircher2016three,cao2020hierarchical,wu2024uav,feng2024fc} efficiently achieve high target completeness, yet assume obstacle-free or fully known environments.
Alternatively, some non-model-based works \cite{song2020online,zhou2021fuel} incorporate online replanning or exploration-oriented approaches to avoid unforeseen obstacles.
However, they primarily prioritize rapidly mapping unknown spaces rather than explicitly enforcing target visibility, thus lacking consideration for comprehensive and clean target observations.
In parallel, visibility objectives are explored in perception-aware planning \cite{zhang2018perception,bartolomei2020perception,lin2025gfm} and aerial object tracking \cite{jeon2020integrated,ji2022elastic,gao2023adaptive}.
The former category favors informative features for state estimation, yet it does not penalize interference from non-target clutter that harms clean target sensing.
The latter imposes sparse line-of-sight (LoS) constraints to keep an object within the FoV, whereas directly transferring such formulations into dense surface coverage is computationally prohibitive at real-time replanning rates.
These limitations prevent existing aerial scanning systems from realizing complete and clean target coverage under newly appearing occluders.

To close this gap, we propose \textbf{FC-Vision}, a real-time visibility-aware replanning framework under the model-based setting for safe, efficient, and occlusion-free target scanning in unknown environments (Fig.\ref{fig:Teaser}).
Given a target structure and its generated nominal scanning path, it proactively adjusts this path to avoid both collisions and target occlusions, while respecting the original plan's full target coverage and efficiency.
To achieve low-latency replanning, we decompose this problem into two levels aligned with completeness and cleanliness in target visibility.
First, a hybrid sampling-and-optimization procedure generates occlusion-free, safe alternative viewpoints that minimally deviate from the nominal set while maintaining equivalent target coverage, and then computes an intent-consistent visiting order by solving a Sequential Ordering Problem to retain efficiency.
Second, we connect consecutive viewpoints using our $\Phi$-A* search in 5-DoF representations (3D position, pitch and yaw), which enables the efficient discrete search in 3D space yet enforces obstacle clearance and FoV-level clean sensing with modest overhead, yielding short connectors with clean target observability along the segment.
The resulting path is finally converted into a minimum-time trajectory, adhering to safety, feasibility, and complete coverage constraints.
Moreover, we develop a plug-in coordination strategy that integrates \textbf{FC-Vision} seamlessly into the existing aerial 3D scanning system without modifying the host architectural stacks.

We benchmark the proposed method against existing representative baselines covering both collision-only and target-aware replanning strategies in diverse challenging simulated scenarios, and further validate its performance through real-world experiments.
\textbf{FC-Vision} consistently boosts target visibility while maintaining comparable flight time under real-time computational budgets.
In particular, it increases target coverage by up to $55.32\%$ and reduces the occlusion ratio by up to $73.17\%$.
Ablations also verify the effectiveness of our key designs.
In summary, the contributions of this paper are:

  1) A real-time replanning framework that explicitly enforces target visibility amid online-emerging obstacles, yet sustaining safety and efficiency, which is augmented via two novel modules: (i) hybrid sampling-and-optimization viewpoint repair, and (ii) $\Phi$-A* for efficient 5-DoF segment search with joint collision and occlusion avoidance. 

  2) A plug-in system integration that allows existing scanning UAVs to adopt \textbf{FC-Vision} in a drop-in manner, without redesigning the underlying pipeline.

  3) Extensive real-world and simulation evaluations in unstructured sites that confirm the practicality and performance of our proposed approach. 
  To the best of our knowledge, this is the first work that supports reliable occlusion-free aerial target structure scanning in unknown environments.
  The source code is available at \url{https://github.com/FC-Family/FC-Vision}.

\vspace{-0.5cm}
\section{Related Work}
\label{sec:related_work}

\subsection{Aerial 3D Scanning Path Planning}

A prevalent paradigm for model-based aerial 3D scanning is to (1) select a set of viewpoints that collectively cover the target surface and (2) compute an efficient feasible tour to visit them.
Representative works \cite{jing2016sampling,bircher2016three,almadhoun2018coverage} optimize surface-coverage objectives together with viewpoint selection and tour construction.
To scale to large and complex scenes, hierarchical planners \cite{cao2020hierarchical,feng2024fc} decompose the target into sub-spaces for efficient global-to-local planning.
Later studies \cite{peng2019adaptive,wu2024uav,viswanathan2025adaptive} further boost viewpoint quality and modeling efficiency, and adapt to target surface morphology via adaptive plane-wise sampling or observability-guided sparse search.
Overall, these planners focus on completeness and efficiency but typically assume static, known environments (or pre-mapped obstacles), which reduces reliability when unexpected occluders appear during execution.

Complementarily, some non-model-based methods \cite{song2020online,zhou2021fuel} adopt frontier-based exploration strategies or navigation-centric replanning to react to newly observed obstacles.
Yet these online decisions are often driven by collision avoidance and maximizing exploration efficiency, while target-centric sensing quality (especially occlusion-free, clean observations for dense surface acquisition) is not enforced explicitly, leaving a gap between \textit{flying safely} and \textit{scanning well}.

\vspace{-0.4cm}
\subsection{Visibility-Aware Planning for UAVs}
Perception-aware planning couples motion with active vision to improve state estimation, \textit{e.g.}, by steering the drone towards feature-rich regions or informative trajectories that reduce uncertainty \cite{zhang2018perception,bartolomei2020perception,lin2025gfm}.
These methods are effective for robust navigation, but their objectives are predominantly robot-centric and typically do not model how non-target occlusions degrade clean target sensing and dense reconstruction.
Visibility constraints are also widely used in aerial tracking, where planners maintain a compact object observable via LoS constraints, 2D visible fans, or differentiable visibility metrics \cite{jeon2020integrated,ji2022elastic,gao2023adaptive}.
These formulations target sparse entities (\textit{e.g.}, a point/box) and are not directly transferable to structure scanning, where dense coverage requires ray-level visibility checks over many surface elements, often significantly expensive for real-time replanning when enforced continuously along a path.

\noindent\textbf{Our Position.}
\textbf{FC-Vision} is related to model-based coverage path planning, but addresses a different online replanning problem for target scanning under unexpected non-target obstacles.
Instead of replacing the upstream coverage planner, our real-time visibility-aware replanning framework refines the nominal scan to actively preclude collisions and occlusions, while sustaining information completeness and operational efficiency.
This capability enables reliable, occlusion-free scanning of target structures in unknown, cluttered environments via only a single-pass flight.

\vspace{-0.4cm}
\section{Problem Formulation}
\label{sec:problem_formulation}

We consider aerial scanning of a known target structure surface $\mathcal{S}$ in an otherwise unknown environment.
Before flight, an upstream model-based coverage planner (\textit{e.g.}, FC-Planner \cite{feng2024fc}) generates an efficient, full-coverage, and collision-free nominal scanning path $\bar{\mathcal{P}}$ with respect to $\mathcal{S}$ assuming that the surrounding space is free except for the target structure.
$\bar{\mathcal{P}}$ is represented as an ordered sequence of 5-DoF camera configurations $q=[\mathbf{p}^{\top},\theta,\psi]^{\top}$ (3D position with pitch and yaw).
The sequence contains: (i) viewpoints observe the target surface through the bounded camera FoV with sensing range limit $r_{\max}$ and (ii) waypoints bridge adjacent viewpoints to make the path dynamically feasible, where each pair of consecutive points is connected by a short segment $\gamma$.
During flight, the UAV constructs an online grid map $\hat{\mathcal{O}}$ to detect non-target obstacles $\Omega$ that do not belong to $\mathcal{S}$ and may invalidate the nominal path by causing collision risk or FoV occlusion.

Thus, we seek to optimize a replanned path $\mathcal{P}$ that preserves the full target-surface coverage and stays close to $\bar{\mathcal{P}}$ for flight efficiency, while maintaining safety and clean sensing:
\vspace{-0.2cm}
\begin{equation}
\small
\begin{aligned}
\min_{\mathcal{P}}\quad &
\mathcal{J}_{\mathrm{dev}}(\mathcal{P},\bar{\mathcal{P}})
+\mathrm{Len}(\mathcal{P}) \\
\text{s.t.}\quad
& \mathcal{C}(\mathcal{P},\mathcal{S}) = \mathcal{C}(\bar{\mathcal{P}},\mathcal{S}),\, \mathrm{dist}\big(\mathcal{P},\hat{\mathcal{O}}\big) \ge d_{\min}, \\
& \mathrm{Occ}\big(\mathcal{P},\Omega\big) = 0,
\end{aligned}
\label{eq:orig_joint}
\end{equation}
where $\mathcal{J}_{\mathrm{dev}}$ penalizes deviation from the nominal path (mainly on viewpoints), $\mathcal{C}(\cdot)$ measures target coverage contributed by viewpoint FoVs, and $\mathrm{Occ}(\cdot)$ indicates FoV-level non-target occlusion.
However, (i) visibility checking depends on discrete ray-geometry interactions over common 3D representations (point clouds, meshes, occupancy), yielding non-smooth and discontinuous constraints; (ii) coverage preservation couples viewpoints and their order through a set-coverage/tour-style combinatorial structure; and (iii) observation cleanliness must be enforced along each segment whenever online map updates.
Together, these factors make the problem highly non-convex, rendering it impractical to solve Eq.\eqref{eq:orig_joint} at real-time rates.

\noindent\textbf{Two-Level Decomposition.}
To enable low-latency replanning, we make Eq.\eqref{eq:orig_joint} tractable by a two-level decomposition in accordance with two visibility criteria: completeness (comprehensive coverage) and cleanliness (fully occlusion-free FoV).

\textit{Level-I: Viewpoint update under coverage and occlusion constraints}
We compute a safe viewpoint set $\mathcal{V}=\{\mathbf{v}_k\}_{k=1}^{K}$ and a visiting order $\pi$ that minimally deviates from the nominal viewpoint set $\bar{\mathcal{V}}$, while preserving nominal coverage and ensuring occlusion-free sensing at the viewpoint-level:
\vspace{-0.2cm}
\begin{equation}
\small
\begin{aligned}
\min_{\mathcal{V},\,\pi}\quad &
\mathcal{D}_{\mathrm{set}}(\mathcal{V},\bar{\mathcal{V}})
\;+\; \sum_{k=1}^{K-1}\mathrm{c}\!\left(\mathbf{v}_{\pi_k},\mathbf{v}_{\pi_{k+1}}\right) \\
\text{s.t.}\quad
& \mathcal{C}(\mathcal{V},\mathcal{S}) = \mathcal{C}(\bar{\mathcal{V}},\mathcal{S}), \, \mathrm{dist}\big(\mathcal{V},\hat{\mathcal{O}}\big) \ge d_{\min}, \\
& \mathrm{Occ}(\mathcal{V},\Omega)=0,
\end{aligned}
\label{eq:level1_ppt}
\end{equation}
where $\mathcal{D}_{\mathrm{set}}$ measures set-to-set discrepancy (\textit{e.g.}, Chamfer distance) and $\mathrm{c}(\cdot,\cdot)$ is the travel cost for tour optimization.

\textit{Level-II: Segment-wise clean-sensing connection in 5-DoF space.}
Given the ordered viewpoints, we connect each consecutive pair with a shortest 5-DoF segment that remains safe and occlusion-free throughout:
\begin{equation}
  \small
\begin{aligned}
\min_{\gamma_k} \quad & \mathrm{Len}(\gamma_k) \\
\text{s.t.}\quad
& \mathrm{dist}\big(\gamma_k,\hat{\mathcal{O}}\big)\ge d_{\min}, \,
\mathrm{Occ}\big(\gamma_k,\Omega\big)=0.
\end{aligned}
\label{eq:level2_ppt}
\end{equation}

\noindent\textbf{Feasibility Boundary.}
Eq.\ref{eq:orig_joint}-\ref{eq:level2_ppt} assume that a collision-safe and clean-sensing repair exists. 
If newly observed obstacles make the passage near the target narrower than the UAV clearance envelope, this feasible set may be empty.
\textbf{FC-Vision} then degrades to collision-free replanning, prioritizing safety while no longer claiming complete clean-sensing coverage.

\vspace{-0.4cm}
\section{Framework Overview}
\label{sec:framework_overview}

\begin{figure}[t]
    \vspace{0.1cm}
	  \centering
    \includegraphics[width=0.97\columnwidth]{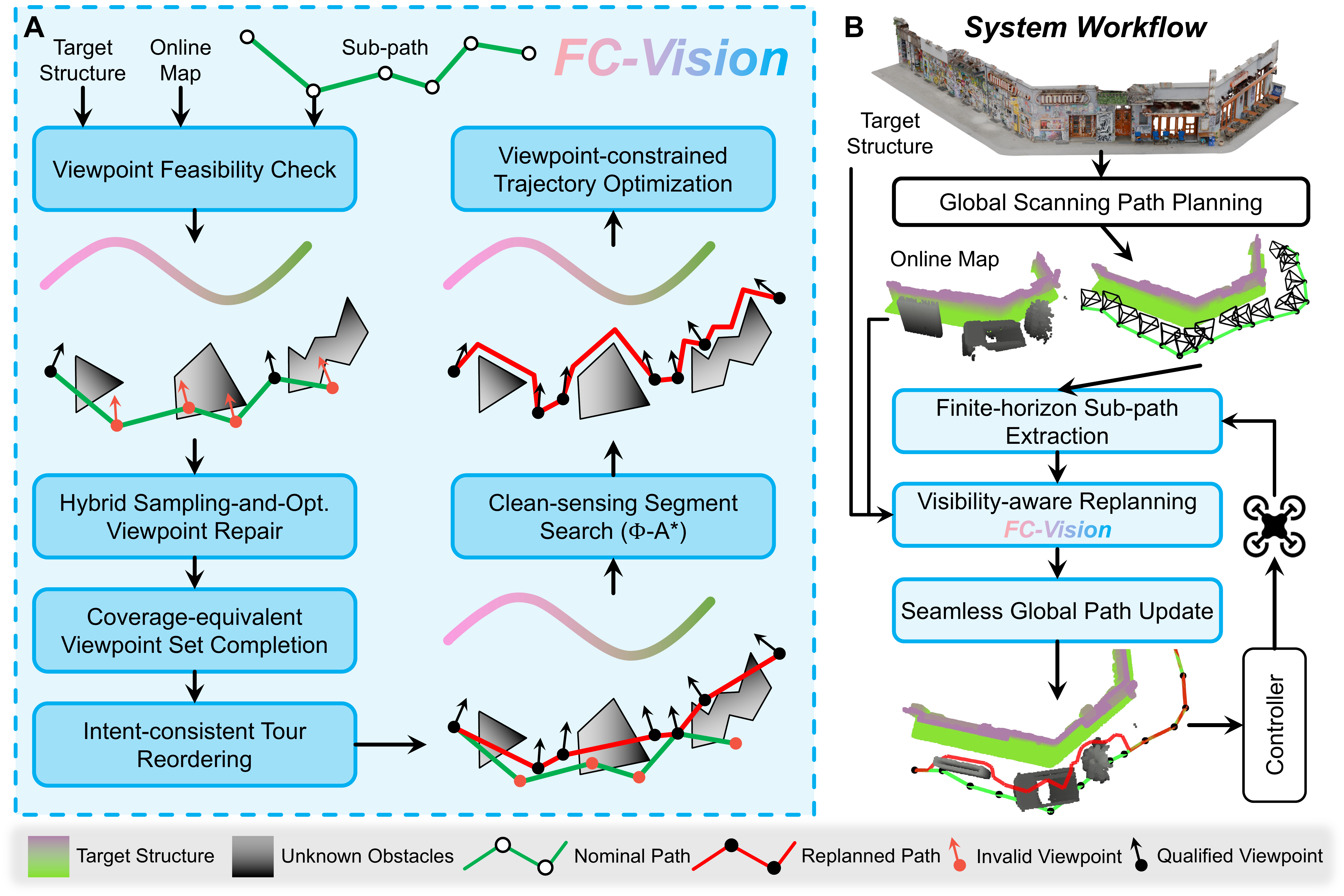}
    \vspace{-0.2cm}
    \caption{\label{fig:Overview} (A) Framework overview \textbf{FC-Vision}. (B) Workflow of aerial scanning system boosted by our visibility-aware replanning (\textbf{Blue}).}
    \vspace{-0.6cm}
\end{figure}

Building on the formulation in Sec.\ref{sec:problem_formulation}, we develop \textbf{FC-Vision} as a low-latency plug-in replanning framework that locally refines the nominal scanning path when flight safety or clean sensing is threatened (Fig.\ref{fig:Overview}).
Upon updating the online map with the latest onboard sensing data, \textbf{FC-Vision} differentiates the non-target obstacles $\Omega$ from the target structure surface $\mathcal{S}$.
Given these identified obstacles, the framework locates and repairs the affected portion of the nominal viewpoint set to preserve full target-surface coverage while enforcing per-viewpoint clearance and occlusion-free observation.
The repaired viewpoints are then reordered to minimize travel cost while respecting the nominal scan progression (Sec.\ref{sub:viewpoint_replanning}).
It next bridges consecutive viewpoint pairs with short, collision-free, and clean-sensing 5-DoF connectors, and assembles them as the replanned path $\mathcal{P}$, which is subsequently converted into a feasible trajectory for execution (Sec.\ref{sub:path_search}).
Finally, \textbf{FC-Vision} serves as a plug-in replanning layer that interfaces with existing UAV
scanning systems: the upstream planner generates $\bar{\mathcal{P}}$ according to the target structure, and our module performs event-triggered real-time path adjustment once newly observed obstacles invalidate the nominal scan, without altering the host pipeline (Sec.\ref{sub:system_integration}).

\vspace{-0.4cm}
\section{Methodology}
\label{sec:methodology}

\vspace{-0.1cm}
\subsection{Occlusion-free Viewpoint Repair with Coverage Preservation and Tour Reordering}
\label{sub:viewpoint_replanning}

To begin, we identify the non-target obstacles $\Omega$ from the online grid map $\hat{\mathcal{O}}$.
Since the target surface is initially known, we voxelize it into $\mathcal{S}=\{\mathbf{e}_j\}_{j=1}^{N}$ before execution and store their voxel indices in a hash set.
Each newly observed occupied voxel is then checked against this hash set and classified into $\Omega$ if it does not correspond to any target voxel.
Next, we detect viewpoints whose intended observations are contaminated by these obstacles from the viewpoint set $\bar{\mathcal{V}}=\{\bar{\mathbf{v}}_i\}$.
For each $\bar{\mathbf{v}}_i$, we collect its intended visible subset $\mathcal{S}_i$ (voxels inside its FoV) and raycast each $\mathbf{e}\in\mathcal{S}_i$ against $\Omega$.
A viewpoint is marked as ``qualified'' if it satisfies both the safety-margin and occlusion-free sensing constraints; otherwise, it is marked as ``invalid'', which are stored in $\bar{\mathcal{V}}_{\mathrm{qual}}$ and $\bar{\mathcal{V}}_{\mathrm{inv}}$, respectively.
To refine $\bar{\mathcal{V}}_{\mathrm{inv}}$, jointly optimizing 5-DoF viewpoint poses and a visiting order forms a non-convex Mixed-Integer Nonlinear Programming (MINLP) with map-dependent visibility constraints, hence we decouple it into (i) viewpoint repair and (ii) tour reordering.

\begin{figure}[t]
    \vspace{0.05cm}
    \centering
    \includegraphics[width=0.7\columnwidth]{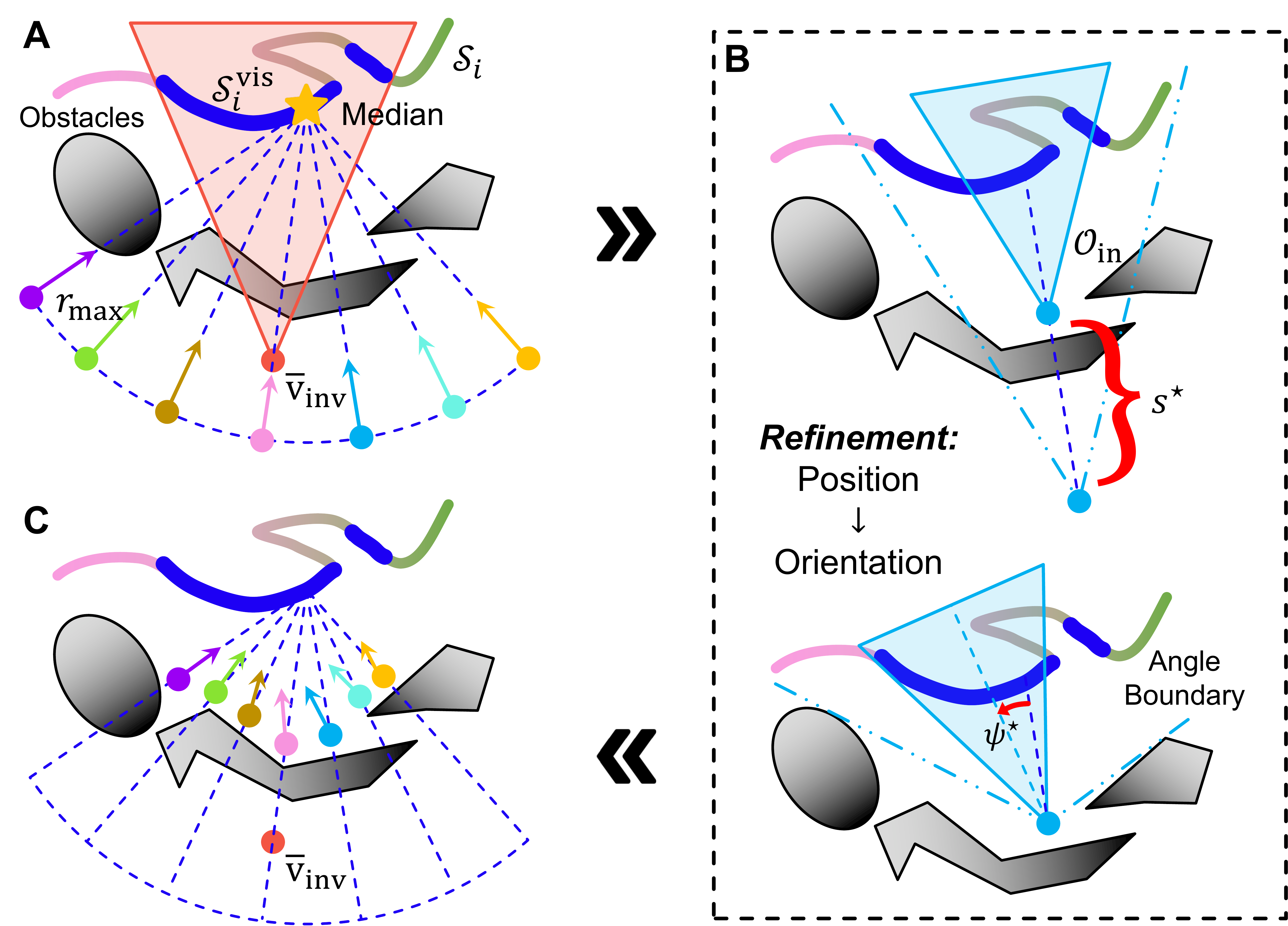}
    \vspace{-0.1cm}
    \caption{\label{fig:ViewpointRepair} Illustration of hybrid sampling-and-optimization viewpoint repair. (A) FoV-truncated spherical sampling. (B) Analytic position optimization and local orientation refinement. (C)  Repaired sampling candidates.}
    \vspace{-0.7cm}
\end{figure}

\noindent\textbf{Hybrid Sampling-and-Optimization Viewpoint Repair.}
Pure sampling-based viewpoint generation in existing coverage or exploration planners would require extensive samples to obtain high-quality repairs, creating a trade-off between solution quality and latency.
Alternatively, direct pose optimization is robust but too slow online due to dense ray-geometry evaluations.
We thus take the best of both: FoV-truncated sampling proposes a small candidate set, and an analytic refinement rapidly improves each candidate under exact frustum constraints (Fig.\ref{fig:ViewpointRepair}).

\textit{(1) FoV-truncated spherical sampling.}
We precompute a reusable direction template in a canonical frame (origin-centered, forward $+x$ axis) and sample angular offsets $(\theta,\psi)$ limited by the horizontal ($\alpha_h$) and vertical ($\alpha_v$) angles of FoV:
\vspace{-0.2cm}
\begin{equation}
  \small
\label{eq:cap_sampling_range}
\theta=n\Delta\theta \in (-\alpha_v,\alpha_v),\
\psi=m\Delta\psi \in (-\alpha_h,\alpha_h).
\end{equation}
\vspace{-0.1cm}
Each sample maps to a unit ray direction $\mathbf{u}(\theta,\psi)=[\cos\psi\cos\theta, \cos\psi\sin\theta, \sin\psi]^\top$ that is generated once and reused.
For an invalid viewpoint $\bar{\mathbf{v}}$, we compute an anchor point $\mathbf{m}(\bar{\mathbf{v}})$ (geometric median of its intended subset) and instantiate candidates on a sphere of radius $r_{\max}$:
\begin{equation}
  \small
\label{eq:sampling_template_new}
\mathbf{p}_i=\mathbf{m}(\bar{\mathbf{v}})+r_{\max}\,\mathbf{R}(\bar{\mathbf{v}})\,\mathbf{u}(\theta_i,\psi_i),
\end{equation}
where $\mathbf{R}(\bar{\mathbf{v}})$ rotates the canonical $+x$ axis to its viewing direction.
This canonical-template reuse avoids per-viewpoint template regeneration and stabilizes candidate quality.

\textit{(2) Analytic position refinement along the viewing ray.}
For a sampled candidate with fixed $(\theta_i,\psi_i)$ and viewing direction $\mathbf{d}$, we optimize only its camera center by translating along $\mathbf{d}$:
\begin{equation}
  \small
\label{eq:ray_param_refine}
\mathbf{p}(s)=\mathbf{p}_i+s\mathbf{d},\quad s\ge 0.
\end{equation}
The FoV frustum in world coordinates is modeled as the intersection of five half-spaces $\mathcal{M} = \{\text{Left}, \text{Right}, \text{Up}, \text{Down}, \text{Far}\}$:
\begin{equation}
  \small
\label{eq:fov_scalar_refine}
\mathcal{F}(\mathbf{p},\theta_i,\psi_i)=\bigcap_{m\in\mathcal{M}}\{\mathbf{x}\in\mathbb{R}^3\mid\mathbf{n}_m^\top \mathbf{x} + h_m(\mathbf{p},\theta_i,\psi_i)\le 0\},
\end{equation}
and the offsets vary affinely with $s$:
\begin{equation}
  \small
\label{eq:h3_update_refine}
h_m^{s}
=
h_m(\mathbf{p}_i,\theta_i,\psi_i)-s\,\mathbf{n}_m^\top\mathbf{d}.
\end{equation}
We first remove self-occluded target voxels under $(\theta_i,\psi_i)$ to obtain $\mathcal{S}_i^{\mathrm{vis}}$.
Coverage at offset $s$ is defined as the fraction of $\mathcal{S}_i^{\mathrm{vis}}$ inside the shifted FoV:
\begin{equation}
  \small
\label{eq:cov_s}
\mathrm{Cov}(s)=\frac{1}{|\mathcal{S}_i^{\mathrm{vis}}|}\sum_{\mathbf{e}\in\mathcal{S}_i^{\mathrm{vis}}}\mathbf{1}\!\Big[\forall m\in\mathcal{M}:\ \mathbf{n}_m^\top \mathbf{e}+h_m^{s}\le 0\Big].
\end{equation}
To enforce clean sensing, let $\mathcal{O}_{\mathrm{in}}$ be obstacle samples inside $\mathcal{F}(\mathbf{p}_i,\theta_i,\psi_i)$ at $s=0$ and define
$\kappa_m(\mathbf{o})\triangleq \mathbf{n}_m^\top \mathbf{o}+h_m(\mathbf{p}_i,\theta_i,\psi_i)$.
The minimum shift to expel $\mathbf{o}$ with safety margin $d_{\min}$ to impose collision avoidance has a closed form:
\begin{equation}
  \small
\label{eq:s_out_refine}
s_{\mathrm{out}}(\mathbf{o})=\min_{m:\ \mathbf{n}_m^\top\mathbf{d}<0}
\frac{d_{\min}-\kappa_m(\mathbf{o})}{-\mathbf{n}_m^\top\mathbf{d}},\ s_{\mathrm{lb}}=\max_{\mathbf{o}\in\mathcal{O}_{\mathrm{in}}} s_{\mathrm{out}}(\mathbf{o}),
\end{equation}
so any $s\ge s_{\mathrm{lb}}$ guarantees an obstacle-free FoV for this ray direction with safe clearance.
We then choose $s$ to maximize coverage under this feasibility bound:
\begin{equation}
  \small
\label{eq:s_opt_problem}
s^\star\in\arg\max_{s\ge s_{\mathrm{lb}}}\ \mathrm{Cov}(s).
\end{equation}
Each voxel $\mathbf{e}\in\mathcal{S}_i^{\mathrm{vis}}$ induces a 1D bound interval $I(\mathbf{e})=[\ell(\mathbf{e}),u(\mathbf{e})]$ from the five affine inequalities $\mathbf{n}_m^\top\mathbf{e}+h_m(\mathbf{p}_i,\theta_i,\psi_i)-s\,\mathbf{n}_m^\top\mathbf{d}\le 0$, hence Eq.\eqref{eq:s_opt_problem} reduces to selecting $s$ that maximizes interval overlap:
\begin{equation}
  \small
\label{eq:max_overlap}
s^\star\in \underset{s\ge s_{\mathrm{lb}}}{\arg\max}\sum_{\mathbf{e}\in\mathcal{S}_i^{\mathrm{vis}}}\mathbf{1}[s\in I(\mathbf{e})],
\end{equation}
solvable in $O(|\mathcal{S}_i^{\mathrm{vis}}|\log|\mathcal{S}_i^{\mathrm{vis}}|)$ via a sweep over interval endpoints \cite{de2008computational}.
The refined candidate is $\mathbf{v}^\star=[\mathbf{p}(s^\star)^\top,\theta_i,\psi_i]^\top$.

\textit{(3) Local pitch and yaw bisection refinement.}
After fixing $\mathbf{p}^\star=\mathbf{p}(s^\star)$, we refine $(\theta_i,\psi_i)$ to further improve visible target coverage within tight admissible bounds derived from the closest obstacle bearings.
Let $\mathcal{O}_{\mathrm{nb}}$ be obstacle samples near the current frustum boundary at $\mathbf{p}^\star$.
Transform a sample $\mathbf{o}\in\mathcal{O}_{\mathrm{nb}}$ into the camera frame under $(\theta_i,\psi_i)$: $\mathbf{r}(\mathbf{o})=\mathbf{R}(\theta_i,\psi_i)^\top(\mathbf{o}-\mathbf{p}^\star)=[x,y,z]^\top,$ and define its horizontal/vertical bearings $\beta_h(\mathbf{o})=\operatorname{arctan}(y,x),\ \beta_v(\mathbf{o})=\operatorname{arctan}(z,x).$
Since the FoV limits correspond to $|\beta_h|\le \alpha_h/2$ and $|\beta_v|\le \alpha_v/2$, the maximum orientation perturbations before $\mathbf{o}$ hits the left/right or up/down plane are given by the angular margins
\begin{equation}
  \small
\label{eq:ang_bound}
\eta_h=\min_{\mathbf{o}\in\mathcal{O}_{\mathrm{nb}}}\Big(|\beta_h(\mathbf{o})|-\frac{\alpha_h}{2}\Big),\,
\eta_v=\min_{\mathbf{o}\in\mathcal{O}_{\mathrm{nb}}}\Big(|\beta_v(\mathbf{o})|-\frac{\alpha_v}{2}\Big),
\end{equation}
yielding $\psi\in[\psi_i-\eta_h,\psi_i+\eta_h]$ and $\theta\in[\theta_i-\eta_v,\theta_i+\eta_v]$.
Within these bounds, we adopt bisection in two dimensions to rapidly find a desired look-at orientation $(\theta_{\mathrm{des}},\psi_{\mathrm{des}})$ that maximizes coverage and does not violate the occlusion-free constraint, producing the final repaired viewpoint $\mathbf{v}^\star=[\mathbf{p}(s^\star)^\top,\theta^\star,\psi^\star]^\top$.

\vspace{0.2cm} 
\noindent\textbf{Coverage-equivalent Viewpoint Set Completion.}
For each invalid viewpoint $\bar{\mathbf{v}}_i$, Steps (1)-(3) yield a candidate pool $\mathcal{Q}_i$ (collision-free, clean FoV).
Jointly selecting a repaired set from $\mathcal{Q}=\cup_i \mathcal{Q}_i$ to preserve global coverage is set-cover-style NP-hard \cite{papadimitriou1998combinatorial}, so we utilize a greedy procedure.
We first pick one replacement per $\bar{\mathbf{v}}_i$:
\begin{equation}
  \small
\label{eq:replace_score}
\mathbf{v}_i^\star=\underset{\mathbf{v}\in\mathcal{Q}_i}{\arg\max}\;\frac{|\mathrm{Cov}(\mathbf{v})\cap\mathcal{S}_i|}{|\mathcal{S}_i|}-\lambda_{\mathrm{d}}\,\|\mathbf{p}(\mathbf{v})-\bar{\mathbf{p}}_i\|_2.
\end{equation}
Replacing all invalid viewpoints yields an initial repaired set $\mathcal{V}_0$, and then a small residual uncovered subset may remain: $\mathcal{U}=\mathcal{S}\setminus \mathrm{Cov}(\mathcal{V}_0)$.
We optionally add a few auxiliary viewpoints to cover $\mathcal{U}$ while staying close to the current set $\mathcal{V}_{\mathrm{cur}}=\mathcal{V}_0\cup\bar{\mathcal{V}}_{\mathrm{qual}}$.
Let $\Delta(\mathbf{v})=\mathrm{Cov}(\mathbf{v})\cap \mathcal{U}$ and $d_{\mathrm{nn}}(\mathbf{v},\mathcal{V}_{\mathrm{cur}})=\underset{\mathbf{u}\in\mathcal{V}_{\mathrm{cur}}}{\min}\|\mathbf{p}(\mathbf{v})-\mathbf{p}(\mathbf{u})\|_2$.
We iteratively select
\begin{equation}
  \small
\label{eq:complete_score}
\mathbf{v}^\star=
\underset{\mathbf{v}\in\mathcal{Q}}{\arg\max}\;
\frac{|\Delta(\mathbf{v})|}{|\mathcal{U}|}
-\lambda_{\mathrm{d}}\,d_{\mathrm{nn}}(\mathbf{v},\mathcal{V}_{\mathrm{cur}}),
\end{equation}
update $\mathcal{V}_{\mathrm{cur}}\leftarrow \mathcal{V}_{\mathrm{cur}}\cup\{\mathbf{v}^\star\}$ and
$\mathcal{U}\leftarrow\mathcal{U}\setminus\Delta(\mathbf{v}^\star)$ until $\mathcal{U}$ is fully covered by all viewpoints in $\mathcal{V}_{\mathrm{cur}}$.

\noindent\textbf{Intent-consistent Tour Reordering.}
After viewpoint repair, we compute an efficient visiting order to retain efficiency but avoid drastically changing the original scan progression.
Consequently, we explicitly preserve the relative order of the originally qualified viewpoints $\bar{\mathcal{V}}_{\mathrm{qual}}=\{\bar{\mathbf{v}}_0,\ldots,\bar{\mathbf{v}}_{M-1}\}$ as precedence anchors, and only allow newly repaired or added viewpoints to be inserted between them. 
This can be formulated as a Sequential Ordering Problem (SOP) \cite{hernadvolgyi2004solving}:
\setcounter{equation}{15}
\begin{equation}
  \small
\begin{aligned}
\label{eq:sop_form}
& \min_{\pi}\ \sum_{k} \mathrm{c}\!\left(\mathbf{v}_{\pi_k},\mathbf{v}_{\pi_{k+1}}\right)
\quad \text{s.t.}\,
\bar{\mathbf{v}}_0 \prec \bar{\mathbf{v}}_1 \prec \cdots \prec \bar{\mathbf{v}}_{M-1}; \\
& \mathrm{c}(\mathbf{v}_i,\mathbf{v}_j)=
\max\left(
\frac{\|\mathbf{p}_i-\mathbf{p}_j\|_2}{v_{\max}},
\frac{|\theta_i-\theta_j|}{\omega_{\max}},
\frac{|\psi_i-\psi_j|}{\omega_{\max}}
\right),
\end{aligned}
\end{equation}
where $\mathrm{c}(\cdot,\cdot)$ is the time-normalized 5-DoF transition cost penalizing both translation and local pitch/yaw changes, while $v_{\max}$ and $\omega_{\max}$ are the translational and angular velocity limits.
The precedence constraints encode the nominal scan progression and prevent large-scale backtracking, while SOP minimizes traversal by inserting repaired viewpoints.

\vspace{-0.4cm}
\subsection{Clean-sensing Segment Search and Trajectory Generation}
\label{sub:path_search}

Given the viewpoint set $\mathcal{V}_{\mathrm{cur}}=\{\mathbf{v}_{\pi_0},\ldots,\mathbf{v}_{\pi_{K-1}}\}$ from Level-I, we bridge each consecutive viewpoints by a short connector that is both collision-free and clean-sensing along the segment.
However, dense ray-level visibility checking is prohibitively expensive, while direct search in discretized 5-DoF space is intractable. 
Thus, we propose a novel $\Phi$-A* search that efficiently finds high-quality clean-sensing segments under critical latency constraints, as shown in Algo.\ref{alg:phi_astar}.

Our key idea is to keep the discrete search in 3D position space while enforcing 5-DoF clean-sensing feasibility through an efficient lifting mapping function $\Phi$.
This avoids the combinatorial explosion of 5D grid search.
During successor expansion, each neighboring 3D position $\mathbf{p}$ is first checked for collision clearance, and then lifted to a 5-DoF camera configuration $q=[\mathbf{p}^\top,\theta,\psi]^\top$ via
\begin{equation}
  \small
\label{eq:phi_lift_final}
\begin{aligned}
    (\theta,\psi) & =\Phi(\mathbf{p}) =(1-\rho)\,(\theta_\text{start},\psi_\text{start})+\rho\,(\theta_\text{goal},\psi_\text{goal}) \\
    \rho & =\frac{\|\mathbf{p}-\mathbf{p}_\text{start}\|_2}{\|\mathbf{p}-\mathbf{p}_\text{start}\|_2+\|\mathbf{p}-\mathbf{p}_\text{goal}\|_2}.
\end{aligned}
\end{equation}
The lifted state is immediately evaluated for FoV cleanliness against non-target obstacles.
When it is occluded, a bounded attitude correction is attempted.
If neither the lifted nor the corrected state satisfies the clean-sensing constraint, the successor is rejected before being inserted into the open set.
Accepted successors are then ordered in the open set by a weighted geometric priority over the camera-position voxel grid:
\begin{equation}
  \small
\label{eq:wastar_final}
f(\mathbf{p}) = g(\mathbf{p}) + \lambda_{\mathrm{heu}}\,\|\mathbf{p}-\mathbf{p}_\text{goal}\|_2,
\end{equation}
where $g$ accumulates Euclidean step costs and $\lambda_{\mathrm{heu}}$ biases the search towards the goal.
Thus, $\Phi$-A* prunes infeasible successors during node expansion, and every resulting 5-DoF connector sample is collision-free and clean-sensing.

\begin{algorithm}[t]
\scriptsize
\caption{$\Phi$-A*: Clean-sensing Segment Search}
\label{alg:phi_astar}
\KwIn{Start/end viewpoints $\mathbf{v}_\text{start},\mathbf{v}_\text{goal}$; Online map $\hat{\mathcal{O}}$; Search step $\Delta_p$; Heuristic weight $\lambda_{\mathrm{heu}}$; Clearance $d_{\min}$}
\KwOut{Clean-sensing connector $\gamma_k$ (or FAIL)}
\textbf{Init:} 
Open set $\mathcal{Y}\leftarrow\{\mathbf{p}_\text{start}\}$; $g(\mathbf{p}_\text{start})\leftarrow 0$; Parent map $\mathrm{par}(\cdot)$; Attitude annotation $\mathrm{att}(\mathbf{p}_\text{start})\leftarrow(\theta_\text{start},\psi_\text{start})$; Visibility cache $\mathrm{cache}\leftarrow \emptyset$\;
\While{$\mathcal{Y}\neq\emptyset$ \textbf{and} within time budget}{
  Pop $\mathbf{p}$ with the smallest $f(\mathbf{p})$ in $\mathcal{Y}$ \tcp*{Eq.\eqref{eq:wastar_final}}
  \If{$\|\mathbf{p}-\mathbf{p}_\text{goal}\|_2 == 0$}{
    Backtrack 3D chain $\{\mathbf{p}_t\}$ using $\mathrm{par}(\cdot)$ and assign $(\theta_t,\psi_t)=\mathrm{att}(\mathbf{p}_t)$\; 
    \Return{$\gamma_k=\{[\mathbf{p}_t^\top,\theta_t,\psi_t]^\top\}$}\;
  }
  \ForEach{neighbor $\mathbf{p}' \in \mathcal{N}(\mathbf{p};\Delta_p)$}{
    \If{$\mathrm{dist}(\mathbf{p}',\hat{\mathcal{O}}) < d_{\min}$}{\textbf{continue}\;}
    $(\theta,\psi)\leftarrow \Phi(\mathbf{p}')$ \tcp*{Eq.\eqref{eq:phi_lift_final}}
    $q\leftarrow[(\mathbf{p}')^\top,\theta,\psi]^\top$\;
    \If{$\mathrm{cache}[\tau(q)]\neq\mathrm{true}$}{
      \If{$\text{FAIL} \leftarrow \mathrm{VisCleanTest}(q,\hat{\mathcal{O}})$}{
        \If{$(\theta,\psi)\leftarrow \mathrm{AttCorrect}(q,\hat{\mathcal{O}})$}{
          $q\leftarrow[(\mathbf{p}')^\top,\theta,\psi]^\top$\;
          $\mathrm{cache}[\tau(q)]\leftarrow \mathrm{true}$\;
        }\Else{
          \textbf{continue} \tcp*{Reject occluded nodes}
        }
      }
    }
    $g_{\mathrm{c}}\leftarrow g(\mathbf{p})+\|\mathbf{p}'-\mathbf{p}\|_2$\;
    \If{$g_{\mathrm{c}} < g(\mathbf{p}')$}{
      $g(\mathbf{p}')\leftarrow g_{\mathrm{c}}$; $\mathrm{par}(\mathbf{p}')\leftarrow \mathbf{p}$; $\mathrm{att}(\mathbf{p}')\leftarrow (\theta,\psi)$\;
      Push or update $\mathbf{p}'$ in $\mathcal{Y}$ with key $f(\mathbf{p}')$\;
    }
  }
}
\Return{FAIL}\;
\end{algorithm}
\setlength{\textfloatsep}{0.0cm}

\noindent\textbf{Bounded Constant-time Occlusion-aware Attitude Correction.}
When the lifted configuration $q$ is occluded, we locally correct $(\theta,\psi)$ under the frustum representation in Eq.\eqref{eq:fov_scalar_refine} rather than discarding it.
To bound the computational overhead, we query a constant-size local voxel set $\mathcal{O}_{\mathrm{loc}}(q)$ and denote $\mathbf{o}^\star$ as the closest sample to the most violated side plane.
The admissible pitch/yaw margins before $\mathbf{o}^\star$ enters the FoV follow Eq.\eqref{eq:ang_bound}, with which we solve a 2D minimum-perturbation adjustment for clean sensing:
\begin{equation}
  \small
\label{eq:min_perturb_repair}
\begin{aligned}
& (\Delta\theta^\star,\Delta\psi^\star)
=\underset{\Delta \theta,\Delta\psi}{\arg\min}\ |\Delta \theta|+|\Delta\psi| \\
& \text{s.t.}\quad \mathrm{Occ}\!\left([\mathbf{p}^\top,\theta+\Delta\theta,\psi+\Delta\psi]^\top,\Omega\right)=0,
\end{aligned}
\end{equation}
Since $\mathbf{o}^\star$ induces a single-boundary feasibility transition with respect to the pitch and yaw axes, the constraint becomes effectively monotone along the interpolation direction, enabling a short bisection to locate the smallest feasible correction.
This procedure is constant-time per expansion: $|\mathcal{O}_{\mathrm{loc}}(q)|$ is bounded by queried voxels, and the bisection runs for at most a fixed $N_{\mathrm{bis}}$ iterations on two axes.
The corrected $(\theta,\psi)$ is stored as a node annotation for path reconstruction.
When the visibility violation cannot be resolved within the bounded correction range, the corresponding spatial node is discarded, forcing the search to seek an alternative route around the occluder.

\noindent\textbf{Visibility Cache for Amortized Acceleration.}
During node expansions, many lifted configurations are repeatedly queried with highly similar $(\mathbf{p},\theta,\psi)$.
We therefore memoize the clean-sensing result using a quantized key $\tau(q)=(\lfloor \mathbf{p}/\Delta_p\rfloor,\lfloor \theta/\Delta_\theta\rfloor,\lfloor \psi/\Delta_\psi\rfloor)$.
Cache hits bypass FoV testing entirely, while cache misses trigger one evaluation and store the result.
This amortizes the cost of visibility checks and substantially improves replanning throughput without weakening the cleanliness constraint.

\noindent\textbf{Trajectory Generation.}
Once all segments are found, we concatenate them with $\mathcal{V}_{\mathrm{cur}}$ to assemble the full repaired path.
This sequence is then converted into a minimum-time and dynamically feasible trajectory using the viewpoint-constrained trajectory optimization \cite{feng2026flyco} satisfying high smoothness, obstacle avoidance, and coverage completeness.

\vspace{-0.5cm}
\subsection{Efficient Plug-in System Integration}
\label{sub:system_integration}

A practical challenge for visibility-aware replanning is deployability: most aerial scanning systems already have a mature stack (global view planning $\rightarrow$ trajectory optimization $\rightarrow$ control), and redesigning the whole pipeline is typically costly.
\textbf{FC-Vision} is thus devised as a plug-in layer that upgrades existing systems to be occlusion-aware through a minimal interface while reusing the host execution modules.
We treat the nominal scanning path $\bar{\mathcal{P}}$ produced by any upstream planner as a reference intent: its ordered sequence of 5-DoF viewpoints fully covers $\mathcal{S}$ and defines the scan progression.
Our algorithm receives $\bar{\mathcal{P}}$ and returns its repaired path $\mathcal{P}$, so that downstream components remain unchanged.
This makes the integration planner-agnostic and allows \textbf{FC-Vision} to be attached to a wide range of scanning stacks.

To meet strict latency, we avoid global re-optimization and operate in a receding-horizon fashion.
At each replanning call, we extract a local sub-path $\bar{\mathcal{P}}_{[i_s,i_e]}$ by accumulating arc-length up to a fixed horizon $\mathcal{H}$ ahead of the current execution.
Our replanning framework then repairs only this window to satisfy collision- and occlusion-free throughout, and we seamlessly splice it back to form the updated scan plan:
\begin{equation}
  \small
\mathcal{P}_{\mathrm{update}}\leftarrow
\bar{\mathcal{P}}_{[0,i_s)}\ \oplus\
\bar{\mathcal{P}}_{[i_s,i_e]}^{\mathrm{replan}}\ \oplus\
\bar{\mathcal{P}}_{(i_e,|\bar{\mathcal{P}}|]}.
\end{equation}
This preserves the upstream intent outside the affected region while reacting promptly to local changes.
This layer is invoked in an event-driven manner whenever non-target obstacles would violate clearance or clean sensing along the current path, and is complemented by a lightweight periodic refresh to remain synchronized with the evolving online map.
Importantly, replanning runs with a strict time budget and returns the best feasible repair found within the budget, ensuring responsiveness in closed-loop flight.
By isolating visibility reasoning into a fast local repair layer, \textbf{FC-Vision} enables drop-in adoption by existing scanners, retaining their mature planning stacks while adding real-time occlusion-aware corrections when the surrounding environment changes.

\vspace{-0.4cm}
\section{Experiments}
\label{sec:exp}

\subsection{Experimental Setup}

\begin{figure}[t]
	\begin{center}
      \vspace{0.1cm}
      \includegraphics[width=0.6\columnwidth]{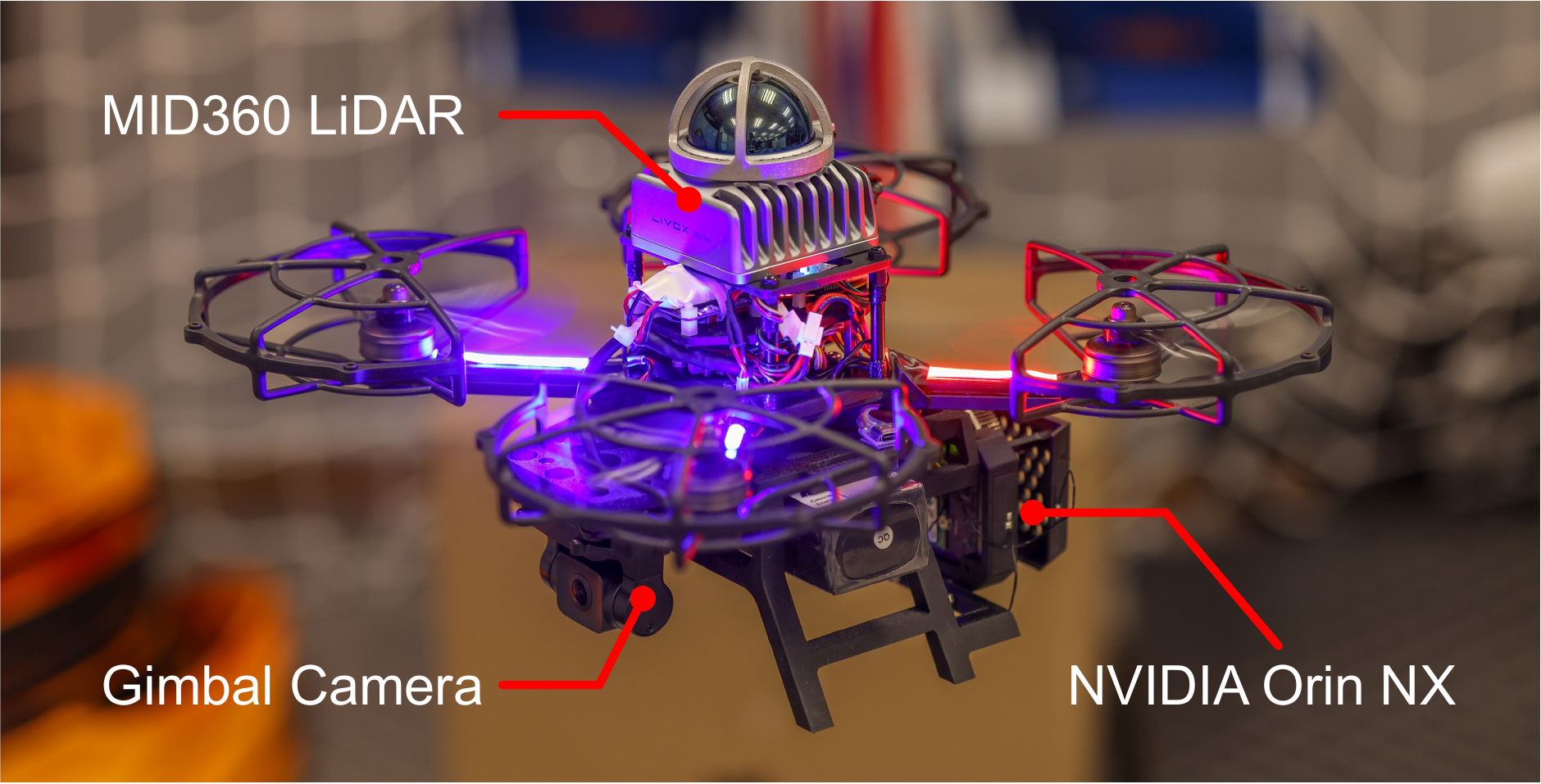}     
      \vspace{-0.5cm}
	\end{center}
   \caption{\label{fig:UAVPlatform} The UAV platform used in real-world scanning flight tests.}
   \vspace{-0.0cm}
\end{figure}

\noindent\textbf{Implementation Details.}
The UAV carries a 3D LiDAR for online mapping and obstacle detection, and a gimbal-mounted RGB camera for surface scanning.
For real-world tests, we customize a quadrotor (Fig.\ref{fig:UAVPlatform}) where the LiDAR and gimbal camera are specified as a Livox MID360 and a SIYI A2 Mini, with all modules running onboard on an NVIDIA Orin NX.
In simulation, AirSim \cite{shah2017airsim} serves as the physics simulator, and the same pipeline is deployed on a desktop workstation (Intel Core i9-12900K CPU). 
To ensure consistent evaluation, we align platform and sensor parameters across both simulation and real flights: the UAV is modeled with a radius of $0.2$ m; the LiDAR measurement range is capped at $15$ m; the camera has a FoV of $[\alpha_h=80^\circ,\alpha_v=65^\circ]$ with the pitch limited to $[-80^\circ,+30^\circ]$; and its effective sensing range is constrained to $r_{\max}=7$ m.
Next, we detail the algorithmic implementation used throughout all experiments.
The online map $\hat{\mathcal{O}}$ is represented by voxels at a resolution of $0.1$ m.
In Sec.\ref{sub:viewpoint_replanning}, the safety margin $d_{\min}$ equals the UAV radius, and $\lambda_{\mathrm{d}}=5.0$ for viewpoint selection.
For the $\Phi$-A* search in Sec.\ref{sub:path_search}, we set the search step $\Delta_p=0.1$ m, heuristic weight $\lambda_{\mathrm{heu}}=10.0$, and the bisection iteration limit $N_{\mathrm{bis}}=10$.
Each replanning cycle uses a horizon $\mathcal{H}=10$ m.
We limit the maximum linear/angular velocity as $1.0$ m/s and $20^\circ$/s in simulation, while $0.5$ m/s and $15^\circ$/s in real-world flights.

\noindent\textbf{Baseline.}
We compare \textbf{FC-Vision} with two baselines.
The first one is \textbf{Colli-Free}, a collision-only replanner following prior obstacle-adaptive aerial scanning methods \cite{song2020online,viswanathan2025adaptive}.
The second is a stronger target-aware baseline, denoted as \textbf{Tar-Aware}, which explicitly accounts for target-surface coverage during replanning.
It constructs an online generalized Voronoi diagram (GVD), samples GVD-guided high-clearance candidate viewpoints around the nominal path, filters FoV-occluded candidates, and selects a compact coverage-preserving viewpoint set via the streaming set cover approach in \cite{song2020online}.
The selected viewpoints are ordered by the same SOP formulation as \textbf{FC-Vision}, connected by conventional collision-free A*, and passed to the same trajectory optimization backend.
For fair comparison, all methods use FC-Planner \cite{feng2024fc} as the upstream planner and are integrated via the proposed plug-in strategy; all other modules are shared, and the only difference lies in the local replanning stage.

\noindent\textbf{Metrics.}
Evaluation focuses on flight efficiency, target visibility, and safety.
Efficiency is measured by the flight time (FT), while visibility by the target coverage rate (CR) that is derived from 3D reconstructions and the occlusion rate (OR), \textit{i.e.}, the fraction of target-occluded frames.
To intuitively reflect overall scanning quality, we define the visibility-adjusted efficiency (VaE):
$\mathrm{VaE} = \mathrm{CR}\cdot(100-\mathrm{OR})/\mathrm{FT}$.
Safety is measured by the success rate (SR) that counts collision-free scan completions.
We also report the replanning computational latency (CL) to assess real-time performance.

\vspace{-0.6cm}
\subsection{Simulation Evaluations}

\vspace{-0.3cm}
\begin{figure}[h]
	\begin{center}
      \includegraphics[width=0.7\columnwidth]{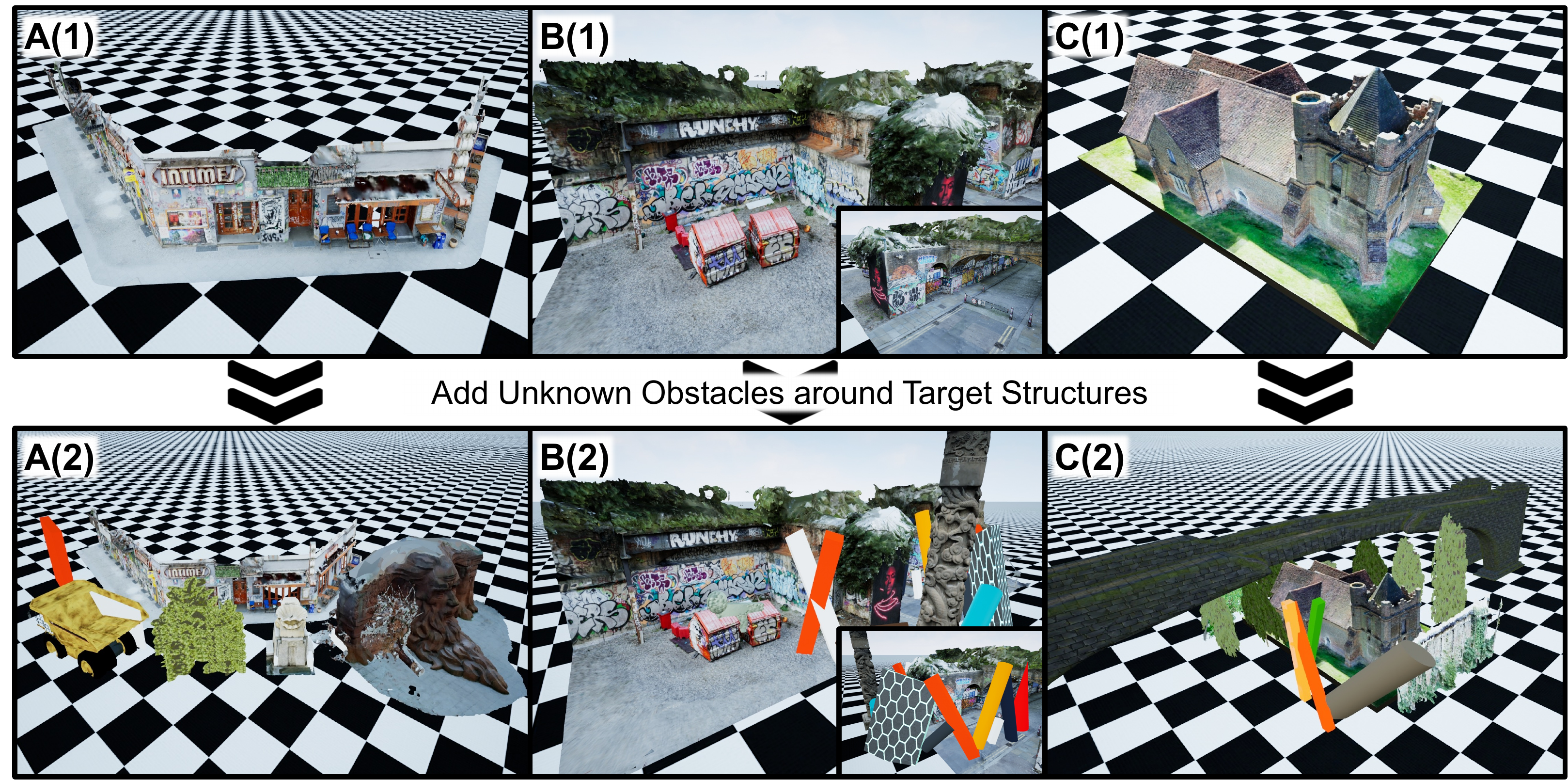}     
      \vspace{-0.6cm}
	\end{center}
   \caption{\label{fig:SimulatedScenarios} Simulated scenarios for benchmark experiments.}
   \vspace{-0.3cm}
\end{figure}

\noindent\textbf{Simulated Environments.}
We benchmark all methods across three challenging simulated scenarios, \textit{Kino Wall (KW)}, \textit{Tunnel (TN)}, and \textit{East Church (EC)}.
Before starting the scan, the UAV only has access to target structure information (top in Fig.\ref{fig:SimulatedScenarios}) and plans a nominal path accordingly.
To emulate unexpected environmental changes, we add randomly complex obstacles around the target (bottom in Fig.\ref{fig:SimulatedScenarios}), and evaluate different methods' capability to handle them.
For each scenario, we run 10 independent trials and report the average results.

\begin{figure*}[t]
	\begin{center}
      \vspace{0.1cm}
      \includegraphics[width=1.95\columnwidth]{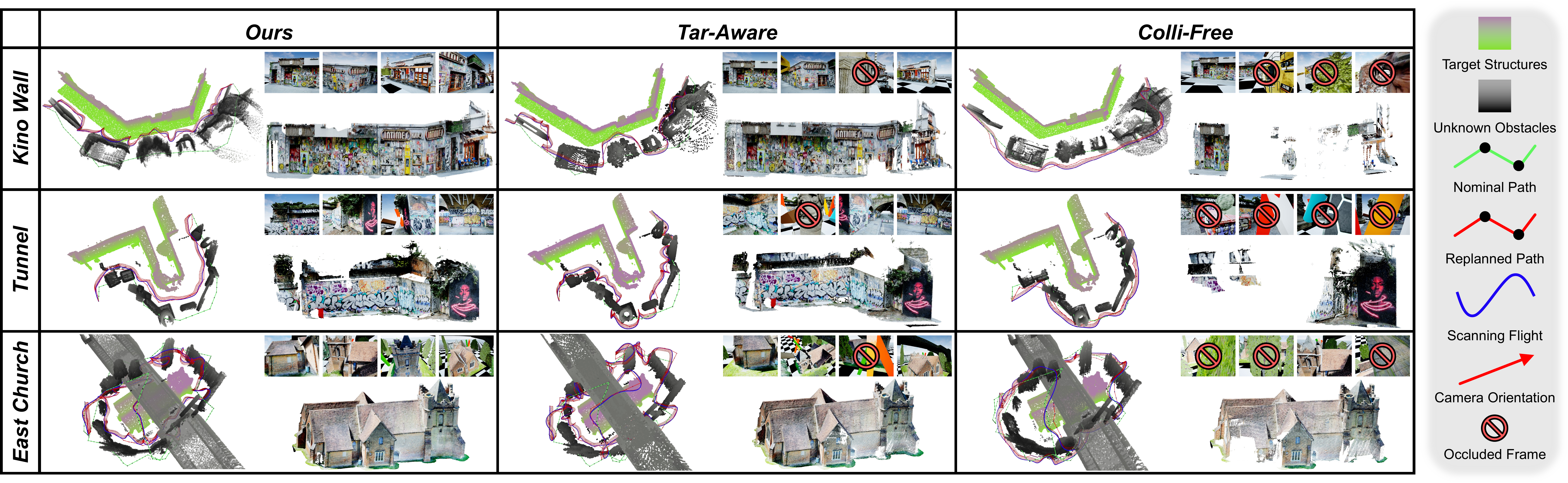}     
      \vspace{-0.6cm}
	\end{center}
   \caption{\label{fig:SimVis} Comparisons on simulated scenarios. Each box includes the flight trajectory, onboard sensing frames, and the snapshot of 3D reconstructions.}
   \vspace{-0.6cm}
\end{figure*}

\noindent\textbf{Comparisons and Analysis.}
Fig.\ref{fig:SimVis} presents the qualitative scanning results, including trajectories, onboard observations, and reconstructed 3D models.
\textbf{Colli-Free} resolves collisions with minimal detours, yet its replanned segments frequently place the FoV behind newly introduced occluders, leading to prolonged occlusions and fragmented reconstructions.
By explicitly considering target-surface coverage, \textbf{Tar-Aware} recovers substantially more complete reconstructions than \textbf{Colli-Free}.
However, since it preserves coverage mainly at the selected viewpoint level, the target can still be occluded along the connecting motion between viewpoints.
In contrast, \textbf{FC-Vision} enforces clean sensing at both the viewpoint and segment levels, thereby producing consistently clear frames and more complete reconstructions across all scenarios.

\vspace{-0.4cm}
\begin{table}[h]
   \renewcommand\arraystretch{1.1}
   \tabcolsep=0.6mm
   \centering
   \scriptsize
   \caption{Simulation benchmark results, reported as averages over 10 runs. \label{tab:SimResults}}
   \vspace{-0.2cm}
   \begin{tabular}{cc|cccccc} 
   \hline
    & Method & FT (s) $\downarrow$ & CR (\%) $\uparrow$ & OR (\%) $\downarrow$ & VaE $\uparrow$ & CL (ms) $\downarrow$ & SR (\%) $\uparrow$ \\
   \hline
   \hline
   \multirow{3}{*}{\rotatebox{90}{\textit{KW}$^{\star}$}} & Colli-Free & \textbf{73.18} & 42.52 & 65.31 & 20.16 & \textbf{19.47} & 100.0 \\ 
    & Tar-Aware & 84.71 & 86.63 & 11.79 & 90.21 & 87.58 & 80.0  \\ 
    & Ours & 79.80 & \textbf{97.84} & \textbf{1.86} & \textbf{120.33} & 23.35 & \textbf{100.0} \\ 
   \hline
   \multirow{3}{*}{\rotatebox{90}{\textit{TN}$^{\dagger}$}} & Colli-Free & \textbf{46.66} & 54.79 & 68.09 & 37.47 & \textbf{18.94} & 100.0 \\ 
    & Tar-Aware & 56.34 & 92.15 & 8.67 & 149.38 & 75.14 & 90.0  \\ 
    & Ours & 49.54 & \textbf{95.51} & \textbf{1.00} & \textbf{190.87} & 27.45 & \textbf{100.0} \\ 
   \hline
   \multirow{3}{*}{\rotatebox{90}{\textit{EC}$^{\ddagger}$}} & Colli-Free & \textbf{149.82} & 75.56 & 53.67 & 23.37 & \textbf{19.33} & 100.0 \\ 
    & Tar-Aware & 195.27 & 85.41 & 13.90 & 37.66 & 96.83 & 70.0  \\
    & Ours & 206.64 & \textbf{94.73} & \textbf{0.72} & \textbf{45.51} & 22.12 & \textbf{100.0} \\ 
   \hline
   \end{tabular}
   \noindent\makebox[\linewidth][l]{\scriptsize$^{\star}$Kino Wall, $^{\dagger}$Tunnel, $^{\ddagger}$East Church.}
   \vspace{-0.3cm}
\end{table}

Table.\ref{tab:SimResults} quantitatively confirms these trends.
Compared with both baselines, FC-Vision achieves near-complete coverage and reduces occlusions to almost zero, while the FT increase remains limited, significantly improving the scanning quality (VaE).
Its clear gain in CR and reduction in OR over \textbf{Tar-Aware} further demonstrates the contribution of our $\Phi$-A* connector search, which prevents FoV contamination along executed segments rather than only at discrete viewpoints.

\vspace{-0.2cm}
\begin{table}[h]
   \renewcommand\arraystretch{1.1}
   \tabcolsep=1.0mm
   \centering
   \scriptsize
   
   \caption{Stage-wise runtime breakdown (ms), averaged over 10 runs. \label{tab:RuntimeBreakdown}}
   \vspace{-0.3cm}
   \begin{tabular}{c|ccc|c|c}
   \hline
   \multirow{2}{*}{Env.} & Viewpoint & Viewpoint Set & Tour Re- & $\Phi$-A* & Total \\
    & Repair & Completion & ordering & Search & Latency \\
   \hline
   \hline
   \textit{KW}& 16.98 & 2.82 & 1.42 & 2.13 & 23.35 \\
   \textit{TN}& 17.86 & 3.73 & 1.79 & 4.07 & 27.45 \\
   \textit{EC}& 15.54 & 2.52 & 1.35 & 2.71 & 22.12 \\
   \hline
   \end{tabular}
   
\end{table}
\vspace{-0.2cm}

Importantly, this benefit does not come from heavier computation: the replanning latency stays low (CL $22$-$27$ ms), meeting real-time requirements.
Table.\ref{tab:RuntimeBreakdown} further breaks down the runtime of \textbf{FC-Vision} across different stages of the replanning pipeline.
In contrast, target-aware operations in \textbf{Tar-Aware} increase CL to $75$-$97$ ms and may contribute to a safety bottleneck, in turn underscoring the advances of our efficient framework design.
Overall, \textbf{FC-Vision} retains the low latency and safety of \textbf{Colli-Free} while achieving stronger target visibility than \textbf{Tar-Aware}.
More details about this experiment can be found in our video.

\begin{figure}[h]
	\begin{center}
      \includegraphics[width=0.7\columnwidth]{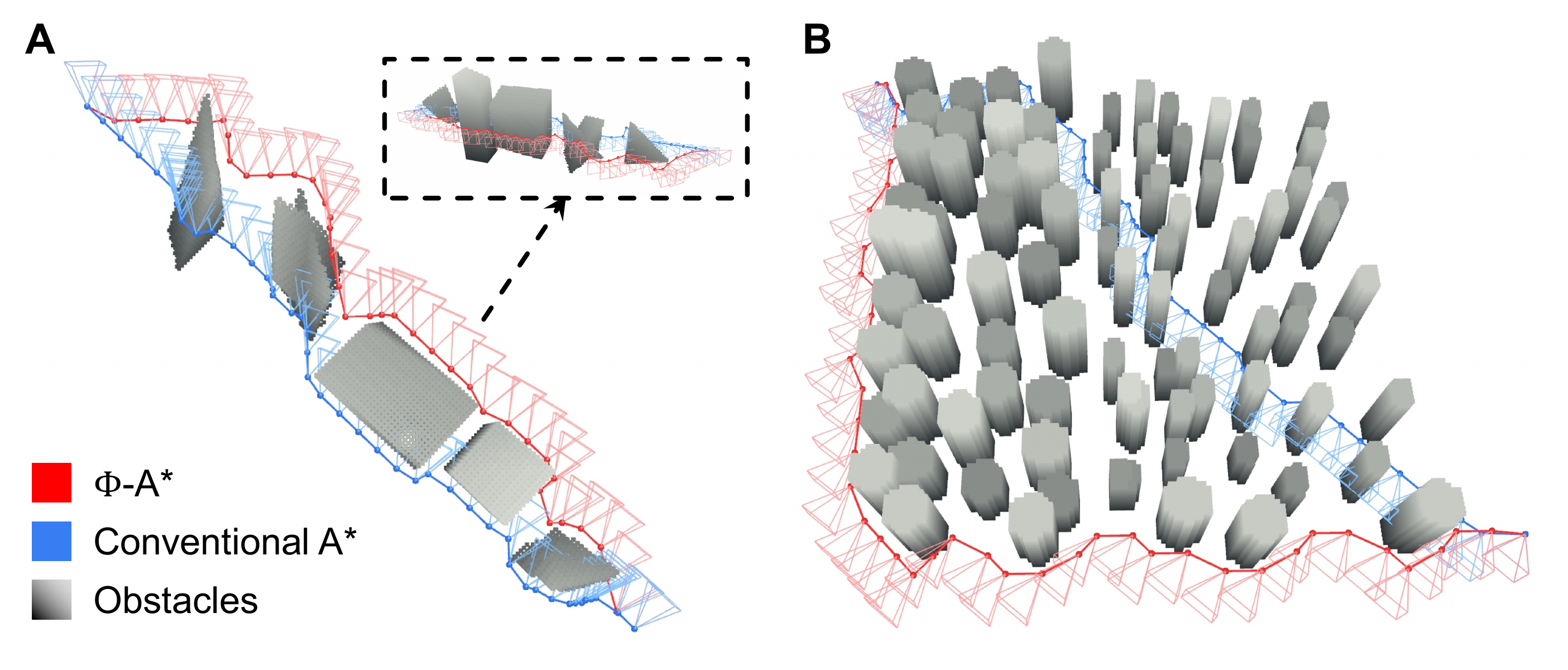}     
      \vspace{-0.6cm}
	\end{center}
   \caption{\label{fig:AstarVis} The qualitative results of $\Phi$-A* vs. conventional A* in (A) corridor and (B) forest cases.}
\end{figure}

\noindent\textbf{Effectiveness of $\Phi$-A*.}
To ablate the contribution of our segment search, Table.\ref{tab:AstarResults} and Fig.\ref{fig:AstarVis} compare $\Phi$-A* with conventional A* on two representative cases (corridor and forest) over 20 repeated runs.
Conventional A* produces shorter connectors but suffers severe visibility degradation (OR $74.78\%$ and $82.62\%$), confirming that collision-free navigation alone does not guarantee clean sensing.
$\Phi$-A* enforces occlusion-free sensing at all accepted connector samples, achieving OR $=0$ in both cases at the cost of slightly longer paths, which is expected since visibility constraints restrict feasible passages.
Moreover, the proposed visibility cache is critical for real-time performance: it reduces the latency of $\Phi$-A* from $3.14$--$3.58$ ms to $1.27$--$1.45$ ms, approaching conventional A* while preserving the same sample-level clean-sensing checks.

\vspace{-0.3cm}
\begin{table}[h]
   \renewcommand\arraystretch{1.1}
   \tabcolsep=1mm
   \centering
   \scriptsize
   \caption{Comparisons of segment search, averaged over 20 runs. \label{tab:AstarResults}}
   \vspace{-0.2cm}
   \begin{tabular}{cc|ccc} 
   \hline
    & Method & CL (ms) $\downarrow$ & Length (m) $\downarrow$ & OR (\%) $\downarrow$ \\
   \hline
   \hline
   \multirow{3}{*}{\rotatebox{90}{{Corridor}}} & Conv. A*$^{\dagger}$ & \textbf{1.154} & \textbf{39.52} & 74.78\\ 
    & Our $\Phi$-A* w/o VC$^{\ddagger}$ & 3.142 & 43.26 & \textbf{0.0}\\ 
    & Our $\Phi$-A* & 1.269 & 43.74 & \textbf{0.0} \\ 
   \hline
   \multirow{3}{*}{\rotatebox{90}{{Forest}}} & Conv. A*$^{\dagger}$ & \textbf{1.286} & \textbf{49.96} & 82.62 \\ 
    & Our $\Phi$-A* w/o VC$^{\ddagger}$ & 3.578 & 64.39 & \textbf{0.0} \\ 
    & Our $\Phi$-A* & 1.453 & 63.85 & \textbf{0.0} \\ 
   \hline
   \vspace{-0.3cm}
   \end{tabular}
   \noindent\makebox[\linewidth][l]{\scriptsize$^{\dagger}$Conventional A*, $^{\ddagger}$$\Phi$-A* without visibility cache acceleration.}
   \vspace{-0.3cm}
\end{table}

\begin{figure*}[t]
	\begin{center}
      \vspace{0.1cm}
      \includegraphics[width=1.95\columnwidth]{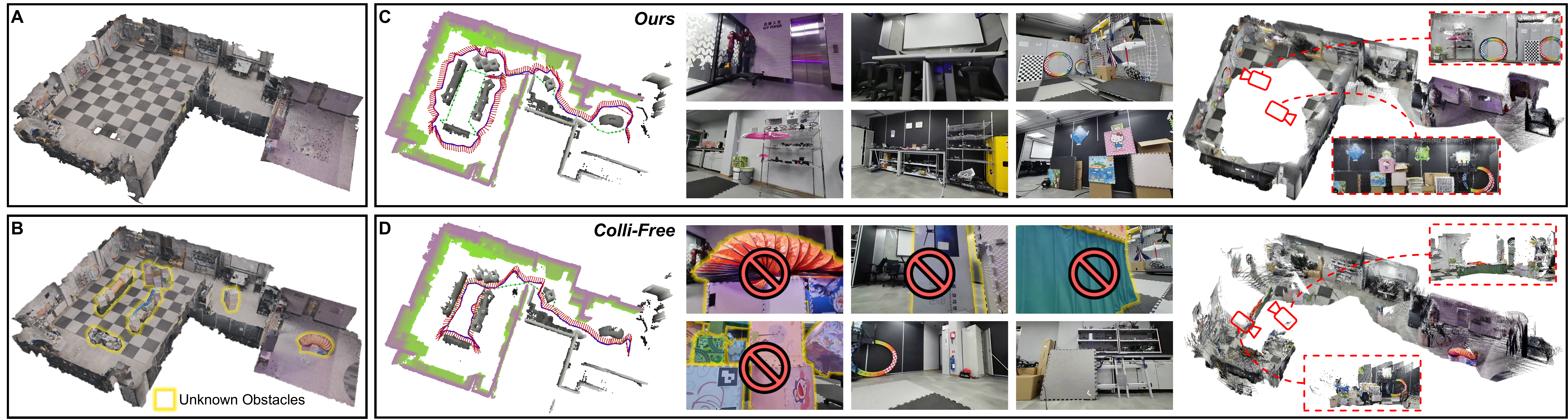}     
      \vspace{-0.4cm}
	\end{center}
   \caption{\label{fig:RealFlightVis} Real-world test results at the \textit{Room} site. (A) Target structure, (B) Environmental changes: newly introduced unknown obstacles, (C)--(D) Flight trajectories, onboard views, and 3D reconstructions by \textbf{FC-Vision} and \textbf{Colli-Free}, respectively.}
   \vspace{-0.6cm}
\end{figure*}

\vspace{-0.5cm}
\subsection{Real-world Flight Tests}

To further assess the deployability of our framework, we conduct fully autonomous scanning flights at two indoor real-world sites, \textit{Rectangular Area (Rect.)} and \textit{Room} (Fig.\ref{fig:Teaser}A and Fig.\ref{fig:RealFlightVis}A).
We compare \textbf{FC-Vision} with the deployed collision-only baseline, \textbf{Colli-Free}, to examine whether visibility-aware replanning indeed enhances scanning quality under physical conditions.
The experimental protocol mirrors simulation: FC-Planner first generates a nominal scanning path for the target structure, after which we insert various objects to alter the environment (Fig.\ref{fig:Teaser}B and Fig.\ref{fig:RealFlightVis}B).

\vspace{-0.3cm}
\begin{table}[t]
   \renewcommand\arraystretch{1.2}
   \tabcolsep=1mm
   \centering
   \scriptsize
   \caption{Statistics of real-world experiments. \label{tab:RealFlightResults}}
   \vspace{-0.3cm}
   \begin{tabular}{cc|ccccc} 
   \hline
    & Method & FT (s) $\downarrow$ & CR (\%) $\uparrow$ & OR (\%) $\downarrow$ & VaE $\uparrow$ & CL (ms) $\downarrow$ \\
   \hline
   \hline
   \multirow{2}{*}{\rotatebox{90}{{Rect.}$^{\star}$}} & Colli-Free & \textbf{20.42} & 84.93 & 73.17 & 111.59 & \textbf{22.39} \\ 
    & Ours & 25.13 & \textbf{98.65} & \textbf{0.0} & \textbf{392.56} & 28.68 \\ 
   \hline
   \multirow{2}{*}{\rotatebox{90}{{Room}}} & Colli-Free & \textbf{133.95} & 61.23 & 48.88 & 23.37 & \textbf{25.16} \\ 
    & Ours & 167.91 & \textbf{97.36} & \textbf{1.19} & \textbf{57.29} & 33.82 \\ 
   \hline
   \vspace{-0.3cm}
   \end{tabular}
\end{table}

\vspace{0.3cm}
Fig.\ref{fig:Teaser} and Fig.\ref{fig:RealFlightVis} show that \textbf{FC-Vision} clearly outperforms the baseline.
Our method proactively routes around online-emerging obstacles and adjusts camera orientation to preserve occlusion-free target observations, resulting in consistently clean frames and substantially more complete surfaces.
Table.\ref{tab:RealFlightResults} corroborates these results: \textbf{FC-Vision} achieves much higher coverage with near-zero occlusion rate, translating into a large-magnitude gain in the overall task quality (VaE), while incurring only a moderate increase in flight time.
Its replanning latency stays low on the edge device, confirming its real-time practicability ($\sim30$ Hz).
These advances stem from our two-level decomposition that turns the expensive optimization under occlusion-free constraint into several efficiently solvable subproblems, achieving a better balance among safety, efficiency, and target visibility in real flights.
Detailed information is provided in our video.

\vspace{-0.4cm}
\section{Conclusion}
\label{sec:conclusion}
We present \textbf{FC-Vision}, a real-time visibility-aware replanning framework empowering occlusion-free and safe aerial scanning of target structure in unknown, cluttered environments. 
It makes target visibility an explicit constraint and achieves low-latency replanning through a two-level decomposition: (1) an efficient occlusion-free viewpoint repair that preserves complete target coverage while minimally deviating from the nominal scan; and (2) a clean-sensing segment search in a 5-DoF representation, which enforces consistent FoV-level cleanliness with collision avoidance along segments with modest overhead. 
We further provide a seamless integration strategy that enables plug-in deployment on existing aerial scanning systems.
Extensive simulation and real-flight results demonstrate the effectiveness, practicality, and efficiency of our approach.
Future work will extend this framework to handle more dynamic scenes and semantic sensing objectives, potentially via end-to-end learning.









\vspace{-0.4cm}
\bibliography{references}

\end{document}